\documentclass[10pt,twocolumn,letterpaper]{article}
\usepackage[pagenumbers]{cvpr}

\usepackage{microtype}
\definecolor{cvprblue}{rgb}{0.21,0.49,0.74}
\usepackage[breaklinks,colorlinks,allcolors=cvprblue]{hyperref}
\hypersetup{
    pdftitle={CoTacWAM: Joint Modeling of Vision and Touch for Robotic Manipulation},
    pdfauthor={Lei Jin, Yiding Ma, Zhuohang Li, Jiachen Liu, Zibo Zhu, Kairui Liu, Chen Gao, Wei Wu, Yong Li}
}

\title{CoTacWAM: Joint Modeling of Vision and Touch for Robotic Manipulation}
\author{
    Lei Jin$^{1,*}$\quad Yiding Ma$^{1,*}$\quad Zhuohang Li$^{1}$\quad
    Jiachen Liu$^{2}$\quad Zibo Zhu$^{1}$\\[3pt]
    Kairui Liu$^{1}$\quad Chen Gao$^{1,\dagger}$\quad Wei Wu$^{3}$\quad
    Yong Li$^{1,\dagger}$\\[5pt]
    {\normalsize $^{1}$Tsinghua University\quad $^{2}$Beihang University\quad $^{3}$Manifold AI}\\[3pt]
    {\small $^{*}$Equal contribution.\quad $^{\dagger}$Corresponding authors.}\\[2pt]
    {\small \href{mailto:chgao96@gmail.com}{\texttt{chgao96@gmail.com}}\quad
    \href{mailto:liyong07@tsinghua.edu.cn}{\texttt{liyong07@tsinghua.edu.cn}}}
}
\date{}

\begin{document}
\maketitle
\begin{abstract}
We present \textbf{CoTacWAM}, a world-action model that learns shared contact semantics between wrist vision and touch, achieving the best observed performance in our nominal manipulation and missing-input evaluations.
The two modalities observe the same local interaction through different signals: wrist images reveal nearby geometry, while touch records physical contact.
Our approach connects their common contact meaning while preserving the evidence unique to each modality, giving the policy a joint description of interaction across available input combinations.
This representation supports predictive action learning alongside head-view scene context and conditions corrections after an independent detector confirms a failed attempt.
Experiments across four real-world tasks demonstrate improved manipulation performance over the evaluated baselines, robustness to missing local observations, and further gains from autonomous retries.
\end{abstract}

\section{Introduction}
\label{sec:intro}

Learning-based manipulation models increasingly translate visual observations and instructions into useful robot behavior~\cite{brohan2023rt1,zitkovich2023rt2}.
Generalist policies extend this capability through diverse robot demonstrations~\cite{octo2024} and pretrained vision--language knowledge~\cite{kim2025openvla,black2025pi0,intelligence2025pi05}.
World-action models further connect action learning with prediction of future sensory changes~\cite{yuan2026fastwam}.
For contact-rich manipulation, useful motion depends on understanding the physical interaction it produces.
Wrist vision and touch offer two observations of this interaction: an image shows the gripper closing around an object, while tactile measurements reveal the resulting contact.
Their signals differ, but both carry evidence about a common event.
Learning what these observations share could help a policy relate visible local changes to physical contact.

Recent policies integrate heterogeneous tactile sensors~\cite{yuan2026ftp1} and exploit visual--tactile relationships through cross-modal fusion and prediction~\cite{heng2025vitacformer,zhang2026restacvla}.
Visuotactile world-action models further connect future sensory prediction with control~\cite{tian2026vtwam,neoteai2026n0twam}.
However, making both streams useful to a policy does not by itself specify a common physical interpretation of their signals.
We therefore ask: \emph{how can wrist vision and touch express common contact meaning, and how can this shared understanding support control?}
Wrist images contribute local geometry, while touch provides direct physical measurements.
A useful joint representation should therefore connect their common meaning while retaining the evidence unique to each modality.
With complete observations, both kinds of information can inform control; with an input missing, the remaining evidence should still support the contact concepts it can reveal.

We introduce \textbf{CoTacWAM}, which organizes wrist vision and touch around shared physical concepts: whether pad--object contact occurs and its coarse strength.
Alongside this common description, the representation preserves the sensory detail needed for manipulation.
By learning this description from paired observations and varying their availability, we form a joint representation that can be read through the same interface across input combinations.
Wrist vision enters through a lightweight trainable interface, while a separate head-view pathway supplies scene context.
We use the learned joint representation to describe current interaction and supervise its future evolution during predictive action learning.
When an attempt fails, an independent detector provides a failure event; the policy combines this context with updated joint observations to generate a correction and resume execution.

We evaluate this approach on four real-world tasks spanning insertion, surface interaction, liquid handling, and fragile-object manipulation.
CoTacWAM achieves 71.3\% mean success without retries, compared with 35.0\% for the strongest baseline.
Representation analyses show that contact semantics are readable from both modalities through a shared head, while the joint representation retains physical interaction detail.
Without wrist vision, the complete system retains 70\% and 75\% success on wiping and tofu grasping, where all three tactile baselines achieve 0\%.
Autonomous retries further improve insertion and wiping success.

Our contributions are threefold:
\begin{itemize}
    \item A joint wrist--tactile representation that establishes shared contact semantics while retaining complementary sensory evidence under varying input availability.
    \item Integration of this representation into predictive action learning and failure-conditioned correction, with head-view scene context supplied separately.
    \item Four-task robot evaluations, representation analyses, and ablations connecting task performance with semantic accessibility, information retention, and temporal learning.
\end{itemize}

\section{Related Work}
\label{sec:related_work}

\subsection{Visuotactile Representation Learning}

Multimodal learning provides a foundation for extracting shared representations from complementary observations~\cite{ngiam2011multimodal}.
Visuotactile learning applies this idea to physical interaction, relating touch to visual observations with different spatial coverage.
Early work uses external scene views for vision--touch prediction~\cite{li2019connecting} and contact-aware multimodal representations~\cite{lee2019making}.
Touch and Go expands paired visual--tactile learning to human-collected interactions in natural environments~\cite{yang2022touchgo}.
SSVTP collects spatially aligned local surface patches~\cite{kerr2023ssvtp}, while UniTouch aligns tactile and visual features across heterogeneous datasets~\cite{yang2024unitouch}.
For policy learning, FTP-1 unifies tactile inputs across sensors~\cite{yuan2026ftp1}, and FuSe connects heterogeneous modalities through language grounding~\cite{jones2025fuse}.
ViTacFormer integrates scene and wrist views with tactile attention and forecasting~\cite{heng2025vitacformer}.

These settings provide different visual contexts for contact.
On robots already equipped with scene and wrist cameras, the wrist view provides local visual evidence without additional sensing hardware or a dedicated setup for aligned surface imaging.
The wrist camera moves with the end effector and observes nearby hand--object geometry, while touch measures the response at the contact interface.
We exploit this physical relationship to learn the contact meaning shared by wrist vision and touch, while preserving complementary sensory evidence.
This local representation and the scene context from a separate head-view pathway jointly inform predictive action learning.

\subsection{World-Action Models for Contact-Rich Manipulation}

Visuomotor policies can generate action sequences directly through action chunking~\cite{zhao2023act} or diffusion~\cite{chi2023diffusion}.
Sensory prediction additionally models the environmental changes associated with those actions.
Visual foresight uses predicted images to plan robot motion~\cite{finn2017foresight}; UniPi generates video plans from language~\cite{du2023unipi}, and UniSim learns interactive visual simulation for policy training~\cite{yang2024unisim}.
RoboScape uses depth and keypoint-dynamics supervision to improve the physical consistency of generated manipulation videos~\cite{shang2025roboscape}.
LongScape improves long-horizon video consistency through action-guided chunking and context-aware expert routing~\cite{shang2025longscape}.
World-action models bring prediction into policy learning.
GR-1 jointly predicts future images and actions after video generative pretraining~\cite{wu2023gr1}, while Fast-WAM co-trains video and action prediction and omits future-video generation at inference~\cite{yuan2026fastwam}.
For contact-rich manipulation, VT-WAM jointly predicts vision, tactile deformation, and actions~\cite{tian2026vtwam}, while $N_0$-TWAM uses force-based tactile representations for predictive control~\cite{neoteai2026n0twam}.
TacWAM incorporates mechanics-aware tactile prediction and temporal interaction history into action learning~\cite{jin2026tacwam}.
These methods make the choice of sensory prediction target consequential for control.
CoTacWAM focuses on how wrist vision and touch jointly describe the physical interaction used for predictive control.
We learn their shared contact semantics before policy training, then use the resulting representation as both current interaction evidence and a future target.

\subsection{Failure Detection and Recovery}

Recovery requires connecting evidence of a failed attempt to an executable correction.
At the task level, SayCan grounds language plans in skill affordances~\cite{ichter2023saycan}, while Inner Monologue feeds execution outcomes back into planning~\cite{huang2023inner}.
REFLECT uses multisensory summaries for failure explanation and corrective planning~\cite{liu2023reflect}, while AHA learns failure detection and explanation from generated trajectories~\cite{duan2024aha}.
To obtain corrective behaviors, FailSafe pairs simulated failures with recovery actions~\cite{lin2026failsafe}, and FLARE combines learned retries with reset skills selected by an online monitor~\cite{zhao2026flare}.

For contact-rich manipulation, corrective actions must also account for the local interaction state following a failure.
Visuotactile regrasping illustrates this connection by predicting the outcomes of candidate grasp adjustments from visual and tactile feedback~\cite{calandra2018regrasp}.
We address this through shared wrist--tactile contact semantics, using the same interaction representation for nominal execution and failure-conditioned correction.
An independent tactile--proprioceptive detector confirms failure, while the policy combines the failure context with continuously updated joint observations to guide adjustment.

\section{Method}
\label{sec:method}

\begin{figure*}[t]
    \centering
    \includegraphics[width=\textwidth]{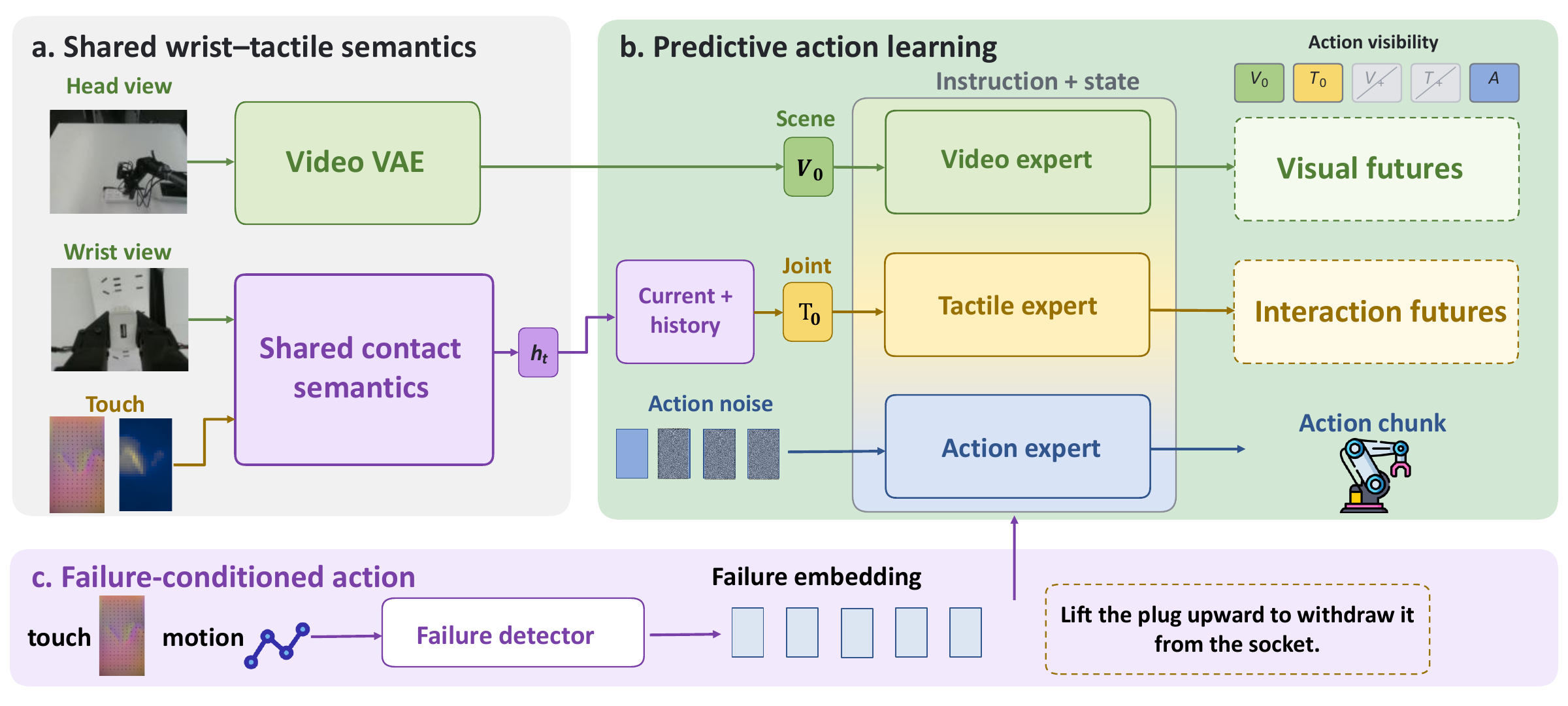}
    \caption{\textbf{CoTacWAM overview.}
    Shared wrist--tactile semantics support predictive action learning and failure-conditioned correction.
    $V/T$ denote scene/interaction tokens, subscripts $0/+$ indicate current/future observations, and $A$ denotes action tokens.}
    \label{fig:method_overview}
\end{figure*}

CoTacWAM connects wrist vision and touch through shared contact semantics, then uses this understanding of local interaction for action learning and failure-conditioned correction (Fig.~\ref{fig:method_overview}).

\subsection{Overview}
\label{sec:method_overview}

At each control step $t$, the robot receives head and wrist images, measurements from two tactile pads, proprioceptive state, and a task instruction.
Using these observations and their available history, it generates an action chunk: a sequence of arm-joint and gripper commands.
The cameras provide different spatial context: the head view covers the scene, while the wrist view observes the region where the gripper interacts with objects.
We retain a separate head-view pathway and introduce wrist vision through a lightweight trainable interface into a local wrist--tactile encoder.
Both pathways supply evidence to action generation throughout execution.

Our central question is how to connect the two local observations so that they express a common interaction state while retaining their complementary information.
We first learn this joint representation, then train action generation together with prediction of its temporal evolution.
When an attempt fails, the same current representation and an independently confirmed failure event condition a corrective response.

\subsection{Learning Shared Wrist--Tactile Semantics}
\label{sec:method_representation}

\begin{figure}[t]
    \centering
    \includegraphics[width=\columnwidth]{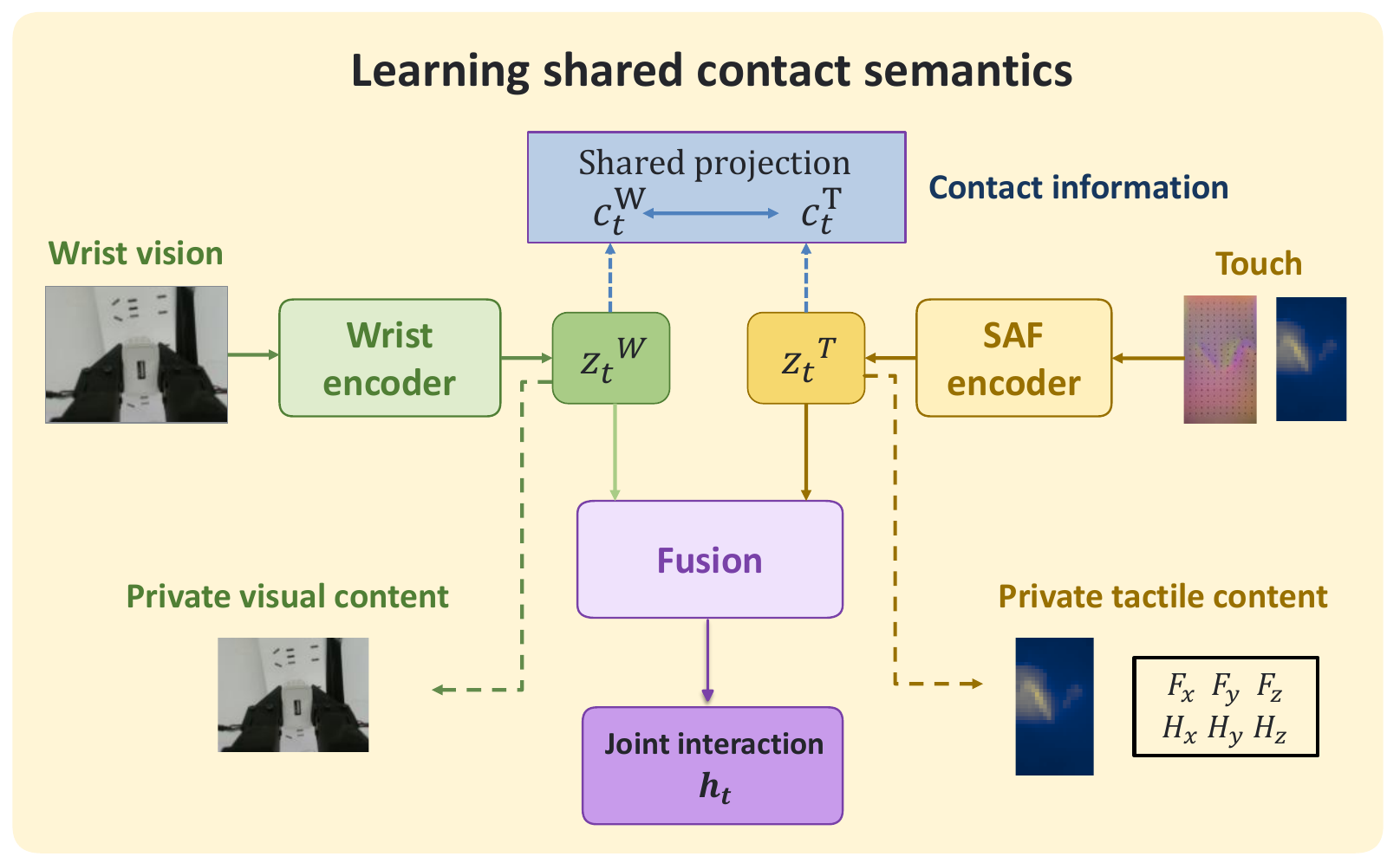}
    \caption{\textbf{Learning shared wrist--tactile semantics while preserving complementary sensory information.}}
    \label{fig:joint_representation}
\end{figure}

We learn a shared semantic reference for wrist vision and touch, while preserving the complementary evidence each modality contributes (Fig.~\ref{fig:joint_representation}).
The resulting joint representation provides a consistent description of contact across different input combinations.

\paragraph{Establishing common contact meaning.}
We use contact presence and coarse strength as interpretable physical concepts that both observations can describe.
Let $z_t^T$ and $z_t^W$ denote the complete tactile and wrist feature vectors; superscripts T and W identify the modalities.
The tactile pathway adapts the Spatially Aligned Fusion (SAF) encoder introduced in TacWAM~\cite{jin2026tacwam}.
SAF fuses spatially corresponding tactile appearance and mechanical signals on each pad before combining the two pads, providing a physically supervised representation of local contact.
Our adaptation uses dense force fields and rectified tactile images and is pretrained through physical reconstruction.
The wrist pathway uses a frozen pretrained Wan2.2 variational autoencoder (VAE)~\cite{wan2025} with a lightweight trainable readout.
Head-view video and rectified tactile images use the same pretrained VAE weights.
Learned projections map the two vectors to semantic features $c_t^T$ and $c_t^W$.
A single prediction head reads either projection to estimate pad--object contact and one of four strength categories: none, light, medium, or high.
This shared head ties both observations to the same physical questions.

Common labels alone leave freedom in how each modality organizes its features.
We further align paired semantic projections using VICReg~\cite{bardes2022vicreg}, encouraging agreement while preventing feature collapse.
The two modalities also provide different evidence for these labels: touch measures a physical response, whereas wrist vision must infer contact from appearance and geometry.
Following the soft-target principle of knowledge distillation~\cite{hinton2015distilling}, we use tactile predictions to teach the wrist branch on trustworthy paired observations.
Together, common supervision defines the semantic reference, alignment connects the feature representations, and tactile-to-visual distillation guides the visual interpretation with tactile evidence.
These constraints act on the semantic projections, leaving the complete vectors available to represent additional information.

\paragraph{Retaining complementary evidence for control.}
Two interactions can share contact and strength categories yet differ in object position or force distribution, calling for different subsequent actions.
A representation organized only around those categories could discard distinctions that matter for manipulation.
We therefore supervise preservation of physical information in the tactile vector and visual information in the wrist vector.
Physical readouts retain local force vectors, per-pad force and torque, and intensity statistics; the visual readout retains a frozen feature summary of the wrist image.
These objectives preserve detail alongside the shared semantic connection.

The complete vectors are then fused by a multilayer perceptron (MLP) into the joint interaction vector $h_t$, which the policy consumes.
Fusion also receives an availability mask $\widetilde m_t$: three binary entries, ordered as left pad, right pad, and wrist view, are one for supplied inputs and zero otherwise.
Physical and visual readouts supervise the fused vector as well, and a separate joint semantic head predicts the same contact and strength concepts.
Thus, information is constrained both before and after fusion, while the shared projections establish the cross-modal correspondence.

Let $\mathcal L_{\mathrm{sem}}$ aggregate contact and strength supervision for the two modalities and joint vector; $\mathcal L_{\mathrm{align}}$ aligns paired semantic features, and $\mathcal L_{\mathrm{distill}}$ transfers tactile predictions to wrist vision.
The term $\mathcal L_{\mathrm{keep}}$ combines the weighted physical and visual preservation losses, while $\lambda_{\mathrm{align}}$ and $\lambda_{\mathrm{distill}}$ are nonnegative scalar weights.
The representation objective combines these roles:
\begin{equation}
\begin{aligned}
    \mathcal L_{\mathrm{repr}}={}&\mathcal L_{\mathrm{sem}}
    +\lambda_{\mathrm{align}}\mathcal L_{\mathrm{align}}\\
    &+\lambda_{\mathrm{distill}}\mathcal L_{\mathrm{distill}}
    +\mathcal L_{\mathrm{keep}}.
\end{aligned}
    \label{eq:representation_objective}
\end{equation}

\paragraph{Maintaining semantic access with incomplete observations.}
A common semantic reference is also useful when only part of the local evidence is available.
Following the principle of modality dropout~\cite{neverova2016moddrop}, we withhold entire modalities and individual pads during training so that the remaining inputs continue to support the same contact description.
For example, deliberately withholding touch can leave a valid recorded contact label with which to supervise a wrist-based estimate.
The explicit mask distinguishes an unavailable pad from a valid pad reporting no contact; a single remaining pad can still contribute measured evidence alongside wrist vision.

Valid recorded semantic labels supervise deliberately masked inputs, while physical and visual preservation losses use the corresponding available source inputs and valid targets.
With complete inputs, both shared cues and complementary detail enter the joint vector; with incomplete inputs, the same interface carries the evidence that remains.
We next use this representation of current interaction to learn its evolution during manipulation.

\subsection{Predictive Action Learning}
\label{sec:method_predictive}

Manipulation depends on how interaction changes over time.
Similar instantaneous contact readings can occur while a grasp is being established or while contact is weakening, making recent observations relevant to the next action.
We encode recent joint representations and proprioceptive history to condition the current interaction anchor, the feature through which the action expert accesses local interaction evidence.
This lets the policy use the current state in the context of its recent evolution.

\paragraph{Learning how interaction evolves.}
To learn beyond the current state, we train action generation alongside prediction of future scene and interaction observations.
We adopt a three-expert predictive framework~\cite{yuan2026fastwam,jin2026tacwam}: Video predicts head-view futures, Tactile predicts future joint wrist--tactile vectors, and Action generates action chunks.
The head-view and interaction prediction tasks cover scene changes and local interaction changes, respectively.

Future observations provide learning targets, while action generation must rely on information available during execution.
Current observations remain clean, unnoised anchors, and an attention mask controls which token groups can exchange information.
Action tokens read the current visual and interaction anchors and other action tokens.
Both the current anchors and Action are blocked from reading future sensory tokens, preventing future observations from entering action generation through either route.
The current interaction anchor and future interaction targets both encode the joint wrist--tactile representation.
This visibility structure allows future sensory supervision to shape predictive learning while actions remain conditioned on observations available at execution.
At execution, the policy generates action chunks using current observations and history and replans as new evidence arrives.

\paragraph{Carrying physical meaning into prediction.}
To preserve contact and physical meaning in future predictions, we apply the frozen physical and semantic readouts learned with the joint encoder to predicted interaction vectors.
Their losses compare predicted local force, force/torque, intensity statistics, and contact/strength with the corresponding valid targets.
Gradients pass through these fixed readouts to the predicted representation.
This carries the semantic reference established between wrist vision and touch into learning how their joint interaction develops.

We combine flow-matching prediction losses~\cite{lipman2023flow} with these readout constraints:
\begin{equation}
\begin{aligned}
    \mathcal L_{\mathrm{policy}}={}&
    \lambda_A\mathcal L_{\mathrm{FM}}^A+\lambda_V\mathcal L_{\mathrm{FM}}^V\\
    &+\lambda_T\mathcal L_{\mathrm{FM}}^T+\mathcal L_{\mathrm{read}}.
\end{aligned}
    \label{eq:policy_objective}
\end{equation}
The flow-matching losses $\mathcal L_{\mathrm{FM}}^A$, $\mathcal L_{\mathrm{FM}}^V$, and $\mathcal L_{\mathrm{FM}}^T$ supervise actions, head-view futures, and joint interaction futures; $\lambda_A$, $\lambda_V$, and $\lambda_T$ are their nonnegative scalar weights.
The term $\mathcal L_{\mathrm{read}}$ is the weighted physical and semantic readout loss on valid future targets.

\subsection{Failure-Conditioned Recovery}
\label{sec:method_recovery}

Failure-conditioned recovery extends the joint interaction representation from nominal execution to corrective actions.
When insertion is blocked or effective wiping contact is lost, recovery combines confirmation of failure with an adjustment informed by the current interaction state.
A lightweight recurrent detector runs on the robot-side controller, independently of the action policy.
It uses tactile force/torque and proprioceptive history to distinguish unsuccessful interaction from normal contact, and initiates correction after sustained failure evidence.

Correction uses two kinds of context: the failed attempt and the interaction currently observed.
We encode the confirmed event together with a snapshot of interaction statistics as an additional condition for the Action expert, accompanied by an adjustment instruction.
The failure context remains fixed during adjustment, while the joint wrist--tactile representation updates with incoming observations.
This gives the learned policy information about the attempt being corrected and the state in which each corrective action is taken.
Corrective demonstrations teach actions such as withdrawing a blocked plug or repositioning the eraser.
Correction ends when the detector confirms task-specific recovery completion: withdrawal while retaining the plug, or restored effective contact for wiping.
The normal instruction then resumes and the extra failure condition is disabled.
Thus, the detector determines when to enter and leave correction, while the policy uses the joint interaction representation to determine how to act.
More details are provided in Appendix~\ref{sec:supp_method}.

\subsection{Training}
\label{sec:method_training}
Training follows three stages, each providing the foundation for the next:
\begin{itemize}
    \item \textbf{Stage 0: SAF Pretraining.} Physical reconstruction trains the adapted SAF tactile backbone to retain local contact information.
    \item \textbf{Stage 1: Joint Wrist--Tactile Representation Learning.} Shared semantic and preservation objectives connect wrist vision and touch while retaining complementary evidence; modality and individual-pad dropout expose the encoder to varying input availability.
    \item \textbf{Stage 2: Predictive Action Learning.} We freeze the joint encoder and video VAE to provide a fixed representation space for current observations and future targets, then train action generation with sensory prediction. Task-specific post-training refines execution and adds reviewed corrective demonstrations for insertion and wiping.
\end{itemize}
The failure detector is trained separately.
More details are provided in Appendix~\ref{sec:supp_recovery}.

\section{Experiments}
\label{sec:experiments}

We first evaluate CoTacWAM on real-world manipulation, then examine how its representation captures shared contact semantics and preserves physical interaction detail.
We further evaluate robustness to missing local observations and the benefits of temporal learning, connecting task performance to the representation and its use in control.

\begin{figure*}[t]
    \centering
    \includegraphics[width=\textwidth]{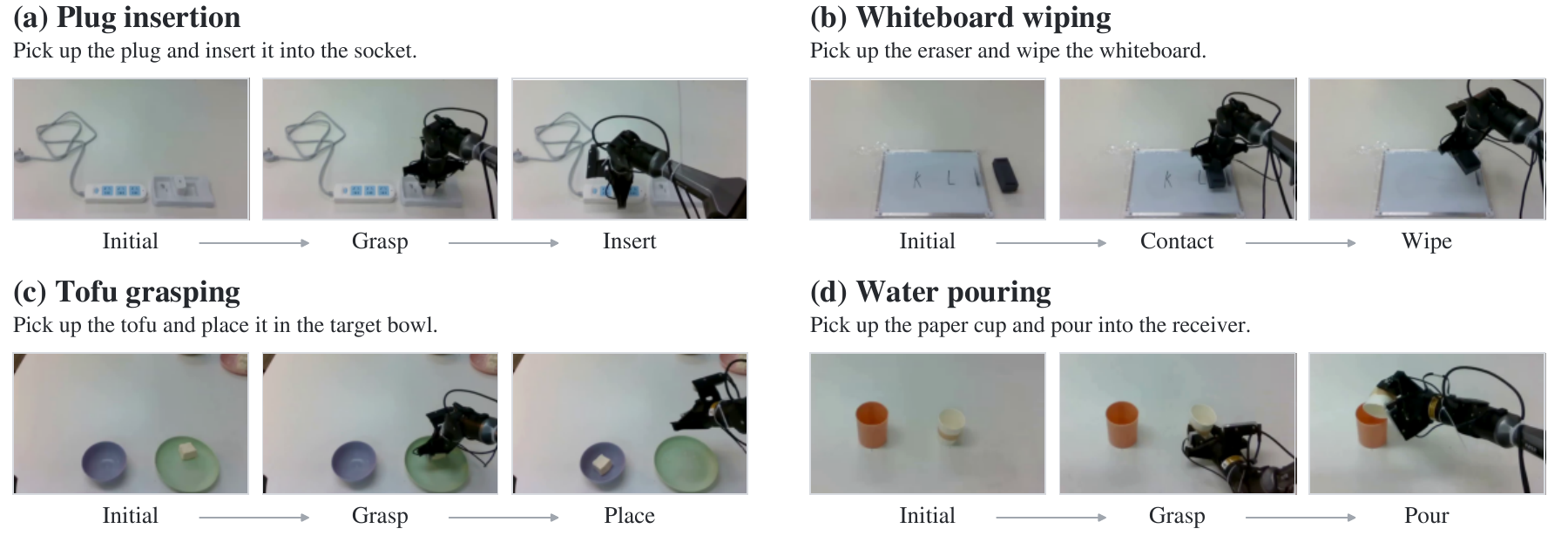}
    \caption{\textbf{Real-world task processes.}}
    \label{fig:task_process}
\end{figure*}

\subsection{Experimental Setup}
\label{sec:exp_setup}

\paragraph{Platform and tasks.}
We use an Agilex Piper manipulator with two gripper-mounted Xense G1-WS tactile sensors, a head camera, and a wrist camera.
The four real-world tasks cover whiteboard wiping, plug insertion, tofu grasping, and water pouring with a paper cup (Fig.~\ref{fig:task_process}).
Insertion emphasizes local alignment, while wiping requires sustained surface interaction.
Tofu grasping requires holding a soft object through transport, and pouring requires controlled reorientation of a held container.
These tasks vary in how local geometry and physical contact inform execution, allowing us to examine the same learned interaction representation across distinct manipulation requirements.
We report final task success over 20 trials per method, condition, and task; detailed evaluation protocols are provided in Appendix~\ref{sec:supp_protocol}.

\paragraph{Training configuration.}
The four task policies share a frozen joint wrist--tactile encoder learned from pooled manipulation data; Sec.~\ref{sec:exp_stage1} examines this same encoder.
Task-specific policies follow the training procedure in Sec.~\ref{sec:method_training}, with the same history and prediction-window settings across tasks.
Training details are provided in Appendix~\ref{sec:supp_training_configuration}.

\paragraph{Baselines.}
We compare two vision-based and three tactile policies.
\begin{itemize}
    \item $\pi_{0.5}$~\cite{intelligence2025pi05} is a generalist vision-language-action policy for instruction-conditioned manipulation.
    \item Fast-WAM~\cite{yuan2026fastwam} co-trains video and action prediction, omitting future-video generation at inference.
    \item $N_0$-VTLA~\cite{neoteai2026n0vtla} integrates latent tactile tokens into a vision-language-action policy.
    \item $N_0$-TWAM~\cite{neoteai2026n0twam} jointly models future vision, contact, and actions using a force-based tactile representation.
    \item FTP-1~\cite{yuan2026ftp1} maps heterogeneous tactile inputs into unified tokens processed by a shared tactile expert.
\end{itemize}
This set covers action policies and predictive models within both visual and tactile families.

\subsection{Real-World Manipulation}
\label{sec:exp_main}

CoTacWAM achieves the highest success rate on all four tasks (Table~\ref{tab:main_results}).
Without retries, CoTacWAM achieves 71.3\% mean success, compared with 35.0\% for the strongest baseline, Fast-WAM, a gain of 36.3 percentage points.
The gains span sustained surface interaction, grasp-and-place, liquid handling, and insertion, demonstrating effective manipulation across different geometric and contact requirements.
This consistent advantage highlights the practical value of a policy that combines local interaction evidence with head-view scene context.

The strongest baseline changes across tasks: $N_0$-TWAM on wiping, $N_0$-VTLA on insertion and tofu, and Fast-WAM on pouring.
CoTacWAM improves over each, showing a consistent advantage across these different task specializations.

Autonomous retries further improve task completion in insertion and wiping.
Allowing retries raises insertion success from 50\% to 80\% and wiping success from 80\% to 100\%.
These final success rates cover complete trials, including the initial attempt and subsequent corrections.
The gains demonstrate the benefit of combining failure confirmation with learned corrective actions, enabling the policy to complete tasks after unsuccessful attempts.

\begin{table*}[t]
    \centering
    \caption{\textbf{Real-world task success (\%).}}
    \label{tab:main_results}
    \normalsize
    \begin{tabular}{lccccc}
        \toprule
        \textbf{Model} & \textbf{Wiping} & \textbf{Insertion} & \textbf{Tofu} & \textbf{Pouring} & \textbf{Avg} \\
        \midrule
        $\pi_{0.5}$ & 30 & 30 & 50 & 25 & 33.8 \\
        Fast-WAM & 20 & 10 & 60 & 50 & 35.0 \\
        $N_0$-VTLA & 15 & 45 & 65 & 10 & 33.8 \\
        $N_0$-TWAM & 55 & 15 & 30 & 10 & 27.5 \\
        FTP-1 & 45 & 15 & 45 & 20 & 31.3 \\
        CoTacWAM (w/o retry) & \textbf{80} & \textbf{50} & \textbf{85} & \textbf{70} & \textbf{71.3} \\
        \midrule
        CoTacWAM (w/ retry) & \textbf{100} & \textbf{80} & --- & --- & --- \\
        \bottomrule
    \end{tabular}
\end{table*}

\subsection{Shared Semantics and Information Retention}
\label{sec:exp_stage1}

\begin{figure*}[t]
    \centering
    \includegraphics[width=0.80\textwidth]{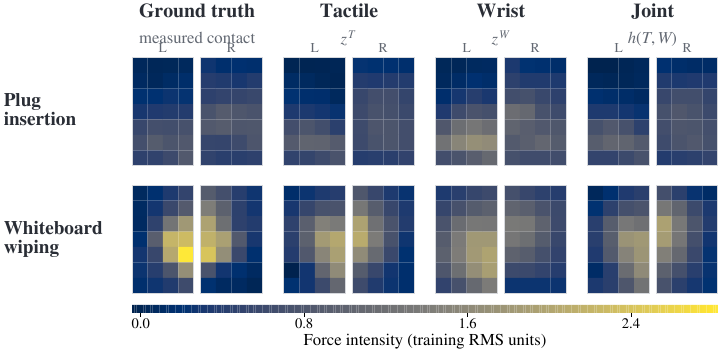}
    \caption{\textbf{Three-way reconstruction.}}
    \label{fig:stage1_threeway_main}
\end{figure*}

We next examine how the learned representation captures shared contact semantics while retaining complementary physical information.
The encoders and trained semantic heads remain frozen throughout these analyses.
We examine contact semantics through the existing readouts and spatial physical information through matched reconstruction probes, using episode-disjoint probe-fitting and evaluation subsets.

\paragraph{Common semantic readout.}
A shared prediction head reads contact and strength information from either wrist vision or touch (Table~\ref{tab:stage1_results}).
Trained on both semantic projections, the head achieves 99.91\% contact accuracy from touch and 92.65\% from wrist vision, compared with 81.37\% for a training-set contact-prior predictor.
Strength accuracy reaches 93.33\% and 79.23\%, respectively.
Both modalities therefore provide access to the same contact concepts, establishing a common semantic interpretation of visual and tactile interaction evidence.
Temporal visualizations in Appendix~\ref{sec:supp_stage1} illustrate how the two readouts follow gripper--object contact in a held-out sequence.

Beyond binary accuracy, the wrist readout achieves a contact Brier score of 0.0646, compared with 0.1522 for the contact prior; the joint readout reaches 0.0087.
This mean squared probability error shows that visual evidence supports informative contact estimates and that the joint representation retains an accurate probabilistic readout.

\paragraph{Retention beyond contact categories.}
Beyond contact categories, the joint representation preserves spatial detail about how forces are distributed across the tactile pads.
We assess this information by fitting a linear ridge probe from each complete vector to a coarse map of local force magnitudes for both pads.
The probes use identical targets, episode splits, feature standardization, and regularization, and are evaluated on the same held-out samples and target scale.

The joint-vector probe achieves 0.1768 MAE, close to the tactile-vector probe's 0.1557 and below task-mean and within-task-shuffled controls (0.4850 and 0.5815).
A task mean captures a typical spatial pattern, while within-task shuffling removes the observation--target correspondence.
The lower error relative to both controls shows that the joint vector retains sample-specific physical information.

Physical retention also extends to force direction.
The native readout predicts signed three-component local force vectors from the joint representation with normalized MAE 0.0451, compared with 0.0328 from the tactile representation.
Together with the intensity-map probe, this readout demonstrates that fusion preserves both directional and spatial aspects of the measured interaction.

Matched reconstructions in Fig.~\ref{fig:stage1_threeway_main} show the spatial structure retained by the joint vector.
In insertion, it preserves the distinct intensity patterns across the two pads.
In wiping, it recovers localized high-intensity regions and the broader intensity variation around them.
These spatial patterns complement the common contact categories by describing how the measured interaction is distributed over the gripper.
Together, the semantic readouts and reconstructions show that the policy's joint input carries both shared contact meaning and detailed physical evidence.
Additional physical readouts and probe details are provided in Appendix~\ref{sec:supp_stage1}.

\begin{table}[!t]
    \centering
    \caption{\textbf{Shared semantics and information retention.}}
    \label{tab:stage1_results}
    \small
    \setlength{\tabcolsep}{4pt}
    \begin{tabular}{@{}lccc@{}}
        \toprule
        Branch & Contact (\%) $\uparrow$ & Strength (\%) $\uparrow$ & Heatmap MAE $\downarrow$ \\
        \midrule
        Tactile & 99.91 & 93.33 & 0.1557 \\
        Wrist & 92.65 & 79.23 & 0.3744 \\
        Joint & 99.06 & 91.37 & 0.1768 \\
        \bottomrule
    \end{tabular}
\end{table}

\begin{table}[!t]
    \centering
    \caption{\textbf{Joint semantic readout with missing inputs (\%).}}
    \label{tab:stage1_missing}
    \small
    \begin{tabular}{lrr}
        \toprule
        Available local inputs & Contact $\uparrow$ & Strength $\uparrow$ \\
        \midrule
        Wrist and touch & 99.06 & 91.37 \\
        Touch only & 99.91 & 93.25 \\
        Wrist only & 92.22 & 78.97 \\
        \bottomrule
    \end{tabular}
\end{table}

\paragraph{Semantic access with incomplete observations.}
The joint interface also preserves access to contact semantics when either local modality is absent.
We keep the encoder and its joint semantic head fixed across complete, touch-only, and wrist-only inputs (Table~\ref{tab:stage1_missing}).
With touch alone, contact and strength accuracies are 99.91\% and 93.25\%; with wrist vision alone, they are 92.22\% and 78.97\%.
Every condition reads the same joint vector used by the policy, preserving a consistent interface as input availability changes.
We next evaluate policy robustness under these conditions.

\subsection{Robustness and Temporal Ablations}
\label{sec:exp_ablations}

\paragraph{Missing local observations.}
CoTacWAM outperforms all three tactile baselines under both missing-input conditions (Table~\ref{tab:missing_modalities}).
We remove wrist vision or touch throughout wiping and tofu rollouts, retaining head vision and proprioception.
All checkpoints are evaluated without additional fine-tuning; our joint representation learning includes modality and single-pad dropout.
Without wrist vision, CoTacWAM retains 70\% and 75\% success, while all three tactile baselines achieve 0\% on both tasks.
Both rates are only ten percentage points below the complete-input results, showing that the policy continues effective manipulation using head vision, touch, and proprioception.

Without touch, CoTacWAM achieves 60\% success on wiping and 55\% on tofu grasping, exceeding the strongest baseline under the same condition by 30 and 20 percentage points, respectively.
Restoring both local modalities further raises success to 80\% and 85\%, demonstrating the value of their complementary evidence.
Together with the semantic readouts, these results support both aims of joint representation learning: retaining useful interaction information when an input is missing and exploiting richer evidence when both are available.

\begin{table}[t]
    \centering
    \caption{\textbf{Missing-modality success (\%).}}
    \label{tab:missing_modalities}
    \small
    \setlength{\tabcolsep}{4pt}
    \begin{tabular}{llrr}
        \toprule
        Model & Missing input & Wiping & Tofu \\
        \midrule
        FTP-1 & Touch & 30 & 35 \\
              & Wrist & 0 & 0 \\
        $N_0$-VTLA & Touch & 10 & 30 \\
                   & Wrist & 0 & 0 \\
        $N_0$-TWAM & Touch & 30 & 25 \\
                   & Wrist & 0 & 0 \\
        CoTacWAM & Touch & \textbf{60} & \textbf{55} \\
                 & Wrist & \textbf{70} & \textbf{75} \\
        \bottomrule
    \end{tabular}
\end{table}

\paragraph{Future supervision and history.}
Both future-interaction supervision and history improve task success, with distinct benefits across the two tasks.
History supplies recent interaction evidence during execution, while future supervision shapes the learned predictive objective.
We retain current inputs and all three experts while removing either future-interaction supervision or historical conditioning (Table~\ref{tab:component_ablations}).
The no-future variant removes future interaction flow-matching and physical/semantic readout losses, while keeping history available.
Future-interaction supervision raises wiping success from 55\% to 80\% and tofu success from 45\% to 85\%, gains of 25 and 40 percentage points.
These improvements show the value of learning how local interaction evolves alongside generating actions from current and past observations.

History likewise raises success from 30\% to 80\% on wiping and from 70\% to 85\% on tofu grasping, gains of 50 and 15 percentage points.
Its larger contribution to wiping highlights the value of recent interaction context for sustained surface contact, while tofu grasping benefits more from future-interaction supervision.
The complete model combines both sources of temporal information to achieve the highest success on both tasks.

\begin{table}[t]
    \centering
    \caption{\textbf{Component ablations: task success (\%).}}
    \label{tab:component_ablations}
    \small
    \begin{tabular}{lrr}
        \toprule
        Variant & Wiping & Tofu \\
        \midrule
        Full model & 80 & 85 \\
        w/o future supervision & 55 & 45 \\
        w/o history & 30 & 70 \\
        \bottomrule
    \end{tabular}
\end{table}

\section{Conclusion}
\label{sec:conclusion}

We presented CoTacWAM, which organizes wrist vision and touch around the contact meaning shared by their observations of an interaction.
Representation analyses show that contact concepts are readable from both modalities and that the joint vector retains physical detail beyond those categories.
In real-world manipulation, the complete model improves task completion, remains useful when a local modality is unavailable, and achieves further gains when retries are enabled.
These findings support learning common physical meaning together with complementary sensory evidence, providing an interaction representation that can serve both predictive action learning and correction after failure.

\label{page:main_text_end}

{\small
\bibliographystyle{ieeenat_fullname}
\bibliography{main}
}

\clearpage
\onecolumn
\appendix
\setcounter{figure}{0}
\setcounter{table}{0}
\setcounter{equation}{0}
\renewcommand{\thefigure}{A\arabic{figure}}
\renewcommand{\thetable}{A\arabic{table}}
\renewcommand{\theequation}{A\arabic{equation}}
\renewcommand{\theHfigure}{appendix.\arabic{figure}}
\renewcommand{\theHtable}{appendix.\arabic{table}}
\renewcommand{\theHequation}{appendix.\arabic{equation}}
\phantomsection
\addcontentsline{toc}{section}{Appendix}
\begin{center}
    {\Large\bfseries Appendix}
\end{center}
This appendix details data processing, training, evaluation protocols, representation analysis, and autonomous recovery.

\section{Data and Real-World Evaluation Protocol}
\label{sec:supp_protocol}

\subsection{Data Collection and Processing}

\paragraph{Collection.}
We collect manipulation demonstrations through teleoperation.
Each recording contains head- and wrist-camera observations, tactile measurements from the two gripper pads, and measured arm-joint and gripper states.
The recordings include nominal task execution and corrective behavior following unsuccessful attempts.

\paragraph{Alignment and quality control.}
Visual, tactile, and robot-state observations are associated using recorded timestamps and source-frame correspondences.
Rectified tactile images and dense force fields are obtained through the sensor processing pipeline.
Validity masks identify missing, stale, or unreliable measurements.
Sensor availability, contact-label validity, and force-target reliability are tracked separately.
Contact and strength supervision is derived from sensor observations as described in Sec.~\ref{sec:supp_label_calibration}; uncertain intervals are excluded from the corresponding objectives.

\paragraph{Representation learning and evaluation.}
Training and evaluation use disjoint source episodes.
Normalization and supervision scales are fitted only on eligible training observations and remain fixed for evaluation.
Training samples are drawn from valid episodes, segments, and temporal windows, with task and contact-state balancing.
Modality and single-pad dropout are applied to available inputs; target validity remains unchanged.

\paragraph{Action supervision.}
Predictive action learning uses reviewed nominal windows and, for insertion and wiping, corrective windows.
Nominal windows can come from successful demonstrations or valid nominal portions of retry recordings.
Deliberately incorrect motions, invalid intervals, and unreviewed actions are excluded from action targets.
History, future sensory targets, and action sequences respect temporal ordering and supervision boundaries.

\subsection{Real-World Evaluation}

\paragraph{Platform.}
The robot is an Agilex Piper manipulator equipped with two Xense G1-WS tactile pads on one gripper, a head camera, and a wrist camera.
Left and right tactile pads refer to the two fingertips of that gripper.
The task policy predicts six arm-joint coordinates and one gripper coordinate.

\paragraph{Trials and reporting.}
Each evaluated method and condition receives 20 trials per task, with a three-minute limit for each complete trial.
A trial succeeds when the robot completes the task-specific goal in Table~\ref{tab:supp_success_criteria} within this limit.
We report the percentage of successful trials; the main comparison additionally reports the unweighted mean across all four tasks.

\begin{table}[ht]
    \centering
    \caption{\textbf{Task completion criteria.} Success is assessed at the complete-task level.}
    \label{tab:supp_success_criteria}
    \begin{tabular}{@{}p{0.20\linewidth}p{0.74\linewidth}@{}}
        \toprule
        \textbf{Task} & \textbf{Success criterion} \\
        \midrule
        Whiteboard wiping & Pick up the eraser and wipe the board clean. \\
        Plug insertion & Pick up the plug and insert it into the socket. \\
        Tofu grasping & Pick up the tofu and place it at the designated location. \\
        Water pouring & Pick up the paper cup and pour water into the receiving container.\\
        \bottomrule
    \end{tabular}
\end{table}

\paragraph{Retry-enabled evaluation.}
For insertion and wiping, we additionally evaluate CoTacWAM with retries enabled.
The robot autonomously triggers retries after unsuccessful attempts, with at most five retries per complete trial.
All attempts and corrective actions share the three-minute trial budget.
Success is final task completion over all 20 trials, including trials that succeed on the initial attempt.

\paragraph{Missing-input evaluation.}
We evaluate CoTacWAM and the three tactile baselines on wiping and tofu grasping after removing either wrist vision or touch throughout each rollout.
Head-camera observations and proprioception remain available; the other local modality is retained.
The checkpoints are evaluated without additional fine-tuning for the missing-input condition.

Implementation of missing inputs and component ablations is described in Sec.~\ref{sec:supp_baselines}.

\section{Implementation and Training Details}
\label{sec:supp_method}

We first detail how wrist vision and touch share a contact-semantic reference while retaining complementary sensory information, then describe how the resulting representation conditions predictive action learning.
The tactile backbone adapts Spatially Aligned Fusion (SAF), introduced in TacWAM~\cite{jin2026tacwam}.

\subsection{Observation Encoding and Availability}

At control step $t$, each action contains six arm-joint coordinates and one gripper coordinate.
Observations carry timestamps and a three-entry binary sensor-validity mask $m_t^{\mathrm{obs}}$, ordered as left tactile pad, right tactile pad, and wrist view.
Each pad supplies a rectified image and a dense three-component force field.
The tactile and wrist feature vectors $z_t^T,z_t^W$ and fused interaction vector $h_t$ each have dimension 64; superscripts T and W identify touch and wrist vision.
Separate learned projection heads produce their 32-dimensional semantic features $c_t^T,c_t^W$.
Both the head-view and joint interaction pathways condition every action chunk.

\paragraph{Pretrained visual encoder.}
We use the pretrained Wan2.2 variational autoencoder (VAE)~\cite{wan2025} from the \nolinkurl{Wan-AI/Wan2.2-TI2V-5B} release for head-view video, wrist RGB, and rectified tactile RGB.
The VAE remains frozen throughout training.
Wrist and tactile images are encoded separately as single-frame clips, while head-view sequences use video encoding.
The wrist pathway adds a trainable spatial adapter and learned query readout to the frozen features; dense force fields use a separate trainable convolutional encoder.

\paragraph{Adapted SAF backbone.}
Our SAF adaptation fuses dense force fields and rectified tactile appearance within each pad, then combines the two pads under physical reconstruction supervision.
Force features and adapted frozen-VAE image features are fused on the force grid, then pooled to a $7\times4$ spatial grid per pad.
Pad-identity and local-position embeddings retain the origin of each spatial feature, and a masked Transformer allows cross-pad interaction before vector compression.
Unavailable pads are excluded as attention keys and their outputs are masked.
Coarse force-vector reconstruction supervises these features, while force and torque readouts constrain a pooled representation during pretraining.
Joint wrist--tactile representation learning transfers this spatial backbone and learns a new tactile readout.
Force targets use native per-pad sensor coordinates.

During joint representation learning, a sampled three-entry binary dropout mask $d_t$ gives input availability
$\widetilde m_t=m_t^{\mathrm{obs}}\odot d_t$.
The product is componentwise and restricts input availability to valid observations.
Missing branches are skipped and represented by fixed placeholders with their masks.
Input availability and target validity remain separate: withholding an otherwise valid observation may leave its semantic supervision available, whereas invalid measurements supply no trustworthy target.
Physical preservation requires valid tactile input and targets; visual preservation requires valid wrist input and targets.
If all local inputs are unavailable, the interaction frame is marked invalid for downstream temporal processing.

A multilayer perceptron (MLP) fuses the complete vectors with their availability indicators:
\begin{equation}
    h_t=\operatorname{MLP}([z_t^T;z_t^W;\widetilde m_t]),
    \label{eq:supp_availability_fusion}
\end{equation}
where semicolons denote concatenation and an entry of one in $\widetilde m_t$ means that the corresponding input is supplied.
The semantic projections define cross-modal supervision; the complete vectors enter fusion.

\subsection{Shared Semantics and Preservation Objectives}

\paragraph{Semantic labels and readouts.}
We omit time indices in the following losses.
The binary target $y\in\{0,1\}$ indicates tactile-pad contact with the grasped object.
The strength class $s\in\{0,1,2,3\}$ bins the larger of the two per-pad resultant-force magnitudes into none, light, medium, and high.
Calibration uses training data only, and contact-label validity is tracked separately from force-amplitude validity.
A single shared prediction head maps either semantic projection to contact probability $p$ and a four-class strength-probability vector $\mathbf q$.
A separate joint head reads the fused vector and predicts the same quantities.
For each readout, semantic supervision is
\begin{equation}
    \ell_{\mathrm{sem}}=\operatorname{BCE}(p,y)+\operatorname{CE}(\mathbf q,s),
    \label{eq:supp_semantic_supervision}
\end{equation}
where BCE and CE are binary and categorical cross-entropy.
Each term is averaged separately over examples with an available readout input and a valid corresponding target; an empty eligible set contributes zero.
The representation loss $\mathcal L_{\mathrm{sem}}$ is the mean of the tactile, wrist, and joint semantic losses.

\paragraph{Alignment and tactile-to-visual distillation.}
The alignment term $\mathcal L_{\mathrm{align}}$ applies VICReg to paired semantic projections, combining feature agreement with variance and covariance regularization.
Alignment and distillation use trustworthy paired examples with the wrist and both pads available.
Let $p^T,\mathbf q^T$ and $p^W,\mathbf q^W$ denote the tactile and wrist semantic predictions, respectively.
The tactile predictions provide stopped-gradient soft targets:
\begin{equation}
    \mathcal L_{\mathrm{distill}}=
    \operatorname{BCE}(p^W,\operatorname{sg}(p^T))
    +D_{\mathrm{KL}}\big(\operatorname{sg}(\mathbf q^T)\,\|\,\mathbf q^W\big),
    \label{eq:supp_semantic_distillation}
\end{equation}
where $D_{\mathrm{KL}}$ denotes Kullback--Leibler divergence between the strength distributions.
The stop-gradient operator $\operatorname{sg}$ fixes the tactile targets within this loss.
The distillation term is averaged over the eligible paired examples.

\paragraph{Physical and visual preservation.}
Tactile readouts retain local force vectors, per-pad force and torque, and intensity statistics, with additional spatial intensity supervision before compression.
The wrist readout preserves a frozen visual-feature summary.
Readouts on the joint vector also preserve these physical and visual targets, so information is constrained before and after fusion.
Regression uses Smooth L1 losses, and $\mathcal L_{\mathrm{keep}}$ denotes their weighted sum.
Preservation losses require valid source inputs and targets.

\subsection{Predictive Objectives and Physical Readouts}

Predictive action learning uses a three-expert framework~\cite{yuan2026fastwam,jin2026tacwam}.
For stream $r\in\{V,T,A\}$, corresponding to head-view video, joint interaction, and actions, let $X^r$ denote its clean target and $\epsilon^r$ Gaussian noise.
With independently sampled noise levels $\sigma_r\in[0,1]$, the flow-matching path and velocity target are
\begin{equation}
    X_{\sigma_r}^r=(1-\sigma_r)X^r+\sigma_r\epsilon^r,
    \qquad u^r=\epsilon^r-X^r.
    \label{eq:supp_flow_path}
\end{equation}
Let $\widehat u_\theta^r$ be the velocity predicted by stream $r$ with trainable parameters $\theta$; its error is $e^r=\widehat u_\theta^r-u^r$.
The masked objective is
\begin{equation}
    \mathcal L_{\mathrm{FM}}^r=
    \mathbb E_{\sigma_r,\epsilon^r}
    \left[w_r(\sigma_r)\|e^r\|_{M^r}^{2}\right],
    \label{eq:supp_masked_flow}
\end{equation}
where $M^r$ selects valid targets and $w_r$ is the scheduler weight.
The masked norm averages over valid components within each window, and windows without valid targets contribute zero.
Invalid targets are cleared before corruption and excluded from the loss.
Current anchors are restored after corruption and excluded from future-prediction losses.
Future visual and interaction targets are detached from frozen encoders and constructed independently of the current input mask.

\paragraph{Attention mask and token visibility.}
The history encoder summarizes joint interaction latents, proprioception, velocity, validity, and timing information and conditions the Tactile expert.
An attention mask separates current observation anchors from future sensory prediction tokens within mixed self-attention.
The current visual and interaction anchors are denoted $V_0,T_0$; zero marks the current point of a prediction window, while $V_+,T_+$ denote future sensory tokens and $A$ denotes action tokens.
The interaction anchor and future interaction targets both encode the joint wrist--tactile representation.
Table~\ref{tab:supp_attention_visibility} specifies the permitted key/value groups for each query group.
Current anchors read only current observations; Action reads the current anchors and action tokens, while each future sensory stream reads its permitted anchors and its own future tokens.
The mask is applied at every layer, so future sensory information cannot reach Action directly or indirectly through the current anchors.
Task language and proprioceptive context are supplied separately from these mixed-attention token groups.

\begin{table}[t]
    \centering
    \caption{\textbf{Attention-mask visibility.}}
    \label{tab:supp_attention_visibility}
    \begin{tabular}{ll}
        \toprule
        Query tokens & Permitted key/value tokens \\
        \midrule
        Current visual anchor $V_0$ & $V_0$ \\
        Future visual tokens $V_+$ & $V_0,V_+$ \\
        Current interaction anchor $T_0$ & $V_0,T_0$ \\
        Future interaction tokens $T_+$ & $V_0,T_0,T_+$ \\
        Action tokens $A$ & $V_0,T_0,A$ \\
        \bottomrule
    \end{tabular}
\end{table}

Four frozen readout groups learned with the joint representation constrain local force vectors, force and torque, intensity statistics, and contact/strength semantics in predicted interaction latents.
For task $k$, interaction latents use a fixed training-set mean $\mu_k$ and scalar RMS $s_k$.
Let $j>0$ denote an offset into the future interaction sequence, and let $\widehat{\bar h}_{t+j}$ and $\widehat h_{t+j}$ denote its predicted clean latent in normalized and native coordinates, respectively.
Using the corrupted interaction target $X_{\sigma_T,j}^T$ and its predicted velocity, we recover these latents before decoding:
\begin{equation}
    \widehat{\bar h}_{t+j}
    =X_{\sigma_T,j}^{T}-\sigma_T\widehat u_{\theta,j}^{T},
    \qquad
    \widehat h_{t+j}=s_k\widehat{\bar h}_{t+j}+\mu_k,
    \quad j>0.
    \label{eq:supp_clean_latent}
\end{equation}
Each readout is compared with its valid measured or derived target.
Its parameters remain fixed while gradients pass through to the predicted latent.

\subsection{Temporal Windows}

Predictive action learning uses task-balanced nominal windows and reviewed corrective windows for insertion and wiping.
Recovery instructions and snapshots follow the causal event sequence.

We use 32 historical observations, 32 action targets, and a current interaction anchor followed by 31 future interaction targets.
The head-camera clip samples nine RGB frames at offsets $0,4,\ldots,32$ before video-VAE encoding.
History and prediction targets respect timestamps, observation validity, and supervision boundaries.

\subsection{Architecture and Optimization Settings}
\label{sec:supp_training_configuration}

Table~\ref{tab:supp_encoder_configuration} summarizes the local encoder dimensions.

\begin{table}[ht]
    \centering
    \caption{\textbf{Local observation encoding.} Image sizes are height $\times$ width; latent shapes are channels $\times$ height $\times$ width.}
    \label{tab:supp_encoder_configuration}
    \begin{tabular}{@{}ll@{}}
        \toprule
        Component & Setting \\
        \midrule
        Wrist RGB / cached latent & $240\times320$ / $48\times15\times20$ \\
        Rectified tactile RGB / latent, per pad & $224\times128$ / $48\times14\times8$ \\
        Dense force field, per pad & $35\times20\times3$ \\
        Pooled force/spatial grid, per pad & $7\times4$ \\
        SAF feature width & 64 \\
        Cross-pad Transformer & 2 layers, 4 attention heads \\
        Tactile / wrist / joint vector & 64 / 64 / 64 dimensions \\
        Tactile / wrist semantic projection & 32 / 32 dimensions \\
        Head-view image for action learning & $192\times256$ \\
        \bottomrule
    \end{tabular}
\end{table}

\paragraph{SAF pretraining and joint representation learning.}
SAF pretraining learns physical spatial features with a base learning rate of $10^{-4}$, BF16 computation, FP32 master weights, weight decay 0.01, and gradient-norm clipping at 1.0.
We pretrain SAF for 90,000 steps and use it to initialize joint representation learning.
Joint representation learning uses AdamW with a global batch of 256 (32 per device on eight devices), no gradient accumulation, a cosine learning-rate schedule, and random seed 42.
The learning rate is $10^{-4}$ for new modules and $10^{-5}$ for the transferred SAF backbone; weight decay and gradient clipping are 0.01 and 1.0.
We train the joint wrist--tactile encoder for 180,000 steps.
The VAE remains frozen, and the tactile backbone, wrist adapter/readout, semantic projections, fusion module, and prediction heads are trainable during joint representation learning.

\paragraph{Availability augmentation.}
Before intersecting with actual sensor validity, joint representation learning samples complete local input, touch-only input, and wrist-only input with probabilities 0.6, 0.2, and 0.2.
Independently, with probability 0.1 it chooses one of the two pads uniformly and withholds it.

\begin{table}[ht]
    \centering
    \caption{\textbf{Representation loss coefficients.} These multiply the corresponding valid-target losses; semantic supervision is averaged across the two modality readouts and the joint readout.}
    \label{tab:supp_representation_weights}
    \begin{tabular}{@{}llr@{}}
        \toprule
        Training component & Objective & Coefficient \\
        \midrule
        SAF pretraining & Local force-vector map / force and torque & 1.0 / 1.0 \\
         & Mean intensity / rectified-image summary & 0.15 / 0.01 \\
        \midrule
        Joint representation & Force and torque / intensity & 1.0 / 1.0 \\
         & Intensity statistics / wrist visual summary & 0.5 / 0.05 \\
         & Shared and joint semantic supervision & 1.0 \\
         & Semantic alignment / tactile-to-wrist distillation & 0.02 / 0.05 \\
         & Tactile-vector / joint-vector local force readout & 1.0 / 0.25 \\
        \bottomrule
    \end{tabular}
\end{table}

VICReg uses agreement, variance, and covariance coefficients of 1.0, 1.0, and 0.1, respectively, with the overall weight in Table~\ref{tab:supp_representation_weights}.
Regression uses Smooth L1 with transition parameter 1.0.

\paragraph{Predictive action learning.}
The task-specific training configurations use AdamW with learning rate $3\times10^{-5}$, betas $(0.9,0.95)$, weight decay 0.01, gradient-norm clipping 1.0, and global batch 256.
The cosine schedule uses a 5\% warmup and a minimum learning rate of 10\% of its peak.
The joint encoder and video VAE remain frozen.
Video, action, and joint-interaction flow-matching losses each have weight 1.0; local-force, force/torque, intensity-statistics, and semantic readout losses each have weight 0.1.
Local-input dropout is used during joint representation learning only.

\subsection{Supervision Scales and Strength Labels}
\label{sec:supp_label_calibration}

Binary contact labels are automatic sensor-response proxies for tactile-pad contact with the grasped object.
The labeling rules use fixed engineering thresholds and retain uncertain intervals as unknown.
At a pad, positive evidence requires a native force-field peak of at least $10^{-3}$, RMS of at least $10^{-4}$, and an RMS/peak ratio of at least 0.03, sustained for two native frames.
Negative evidence requires a peak at most $10^{-4}$, RMS at most $10^{-5}$, and force-resultant magnitude at most 0.01 for three frames; available recorded force evidence must also support absence.
Low-response closed-gripper intervals, inconsistent force readings, and invalid observations are marked unknown.
Gaps exceeding 0.1 seconds reset confirmation.
A frame is positive if either valid pad is positive, negative only if both pads are valid and negative, and unknown otherwise.
Unknown labels are excluded from semantic supervision and evaluation.

\paragraph{Strength calibration.}
Strength labels are generated separately from trusted contiguous force/torque sequences, using the larger of the two pad force-resultant magnitudes.
A training-fitted hysteresis gate determines the zero-strength state; positive magnitudes are divided into light, medium, and high classes using training quantiles.
This auxiliary gate does not relabel the binary contact target.

The gate uses the 85th training percentile with an onset floor of 1.0, an offset threshold at half the onset, and a two-frame minimum run length for on/off states.
Positive strength boundaries use the one-third and two-thirds quantiles.
All thresholds remain fixed during validation.
Contact and strength targets have separate validity masks.

Local force-vector targets pool signed vectors to $7\times4$ cells per pad, whereas intensity targets first take each native vector's magnitude and then pool spatially.
Both use a fixed training-fitted RMS scale without centering or clipping, preserving zero force.
Force/torque readouts instead use componentwise training means and standard deviations without clipping.
Force quantities use native sensor units.

\section{Baseline and Ablation Details}
\label{sec:supp_baselines}

\subsection{Baseline Methods}

The comparison covers vision-based action generation, visual world modeling, and tactile-conditioned policies with different ways of connecting contact observations to actions.
All are evaluated by complete-task success under the protocol in Sec.~\ref{sec:supp_protocol}.

\paragraph{$\pi_{0.5}$.}
This vision--language--action model combines a pretrained vision--language backbone with a flow-matching action expert~\cite{intelligence2025pi05}.
Its training recipe combines robot demonstrations with vision--language data to transfer semantic knowledge to manipulation, while continuous action generation supports fine-grained control.
Given camera observations, a task instruction, and the robot state, the action expert generates a chunk of future controls.
Our evaluation uses the instruction-conditioned manipulation policy with head and wrist images and proprioception, without tactile input.

\paragraph{Fast-WAM.}
Fast-WAM learns from future video during training while generating actions directly from current observations at inference~\cite{yuan2026fastwam}.
It combines a pretrained Wan video diffusion backbone with an action Transformer through shared attention.
Both future-video prediction and action prediction access the clean current-frame tokens, but the attention mask prevents action tokens from accessing future-video tokens.
Consequently, inference encodes the current visual context once and denoises the action chunk without generating a future video.
Our implementation supplies head and wrist RGB images together with proprioception and the task instruction.

\paragraph{$N_0$-VTLA.}
$N_0$-VTLA adds a predictive tactile pathway to a vision--language--action backbone~\cite{neoteai2026n0vtla}.
A lightweight predictor combines tactile features with the visual and language context to estimate latent tokens representing the contact change expected over the upcoming action chunk.
These tokens condition the flow-matching action expert directly, rather than entering the vision--language prefix as additional image tokens.
The original training recipe first grounds this predictor in future tactile targets, then trains the action expert to use its output before joint optimization.
In our wiping and tofu evaluations, the policy receives dense tactile force fields from both pads, including current readings and episode-initial references, alongside camera observations and proprioception.

\paragraph{$N_0$-TWAM.}
$N_0$-TWAM models future vision, touch, and actions with separate video, tactile, and action experts connected through shared attention~\cite{neoteai2026n0twam}.
Its causal generation schedule first predicts future scene and contact observations, then conditions action generation on these predictions and the observation history.
The published model also introduces an observed-touch pathway that supplies current tactile feedback to the action expert.
In our evaluation, the deployed model takes multi-view RGB and two rectified optical tactile streams, with observed history used to update its streaming state between action chunks.

\paragraph{FTP-1.}
FTP-1 learns a shared tactile policy across heterogeneous sensors using a morphology-aware tactile token space~\cite{yuan2026ftp1}.
Sensor-specific encoders map image, array, or state measurements into tokens associated with functional contact regions, allowing the tactile expert to share knowledge across sensor types.
Built on $\pi_{0.5}$, it uses separate vision--language, tactile, and action experts: the first two process their own streams, while action queries attend to both when generating controls through flow matching.
Our implementation uses current rectified optical images from the two fingertip pads, preserving their distinct functional-area identities, together with head and wrist images and proprioception.

For a given method and task, missing-input evaluation uses the same trained parameters as its complete-input condition, without additional fine-tuning.
CoTacWAM uses both dense force and rectified tactile appearance in its tactile pathway; the baseline input-removal procedures are specified below.

\subsection{Removing Local Observations}

The missing-input intervention removes information at the policy-input boundary while retaining head-camera observations and proprioception.
It is applied throughout a new rollout, including any reference frames or cached observations used by the policy.

\begin{table}[ht]
    \centering
    \caption{\textbf{Missing-input implementation for wiping and tofu.} Input removal follows each deployed model interface.}
    \label{tab:supp_missing_implementation}
    \begin{tabular}{@{}p{0.13\linewidth}p{0.37\linewidth}p{0.43\linewidth}@{}}
        \toprule
        Method & Wrist removed & Both tactile pads removed \\
        \midrule
        FTP-1 & Omit the wrist input. & Omit tactile inputs and use the native vision--action two-expert inference path, retaining the learned parameters. \\
        $N_0$-VTLA & Omit the wrist source and use the native input mask. & Omit both force sources; mark both current and episode-initial tactile references unavailable. \\
        $N_0$-TWAM & Zero the wrist RGB streams, including historical grounding frames. & Zero both optical tactile streams, including historical grounding frames; retain fixed input keys and shapes. \\
        CoTacWAM & Remove wrist evidence and mark it unavailable in the joint encoder. & Remove both tactile pads and mark them unavailable in the joint encoder. \\
        \bottomrule
    \end{tabular}
\end{table}

The fixed-stream $N_0$-TWAM interface encodes the zero images through its original encoders, which can produce nonzero features.
Each condition starts with fresh episode state and applies the same input removal to current observations, references, and cached history.

CoTacWAM uses modality and single-pad dropout during joint representation learning (Sec.~\ref{sec:supp_training_configuration}).

\subsection{Temporal Component Ablations}

Both internal ablations use wiping and tofu, with 20 complete-task trials per condition and task.
\emph{No future supervision} disables the future joint-interaction flow-matching loss and the physical/semantic readout losses on predicted interaction states.
It retains current interaction inputs, temporal history, all three experts, and the video/action objectives.
\emph{No history} removes the temporal-memory input while retaining current observations and future-interaction learning.

\section{Joint Wrist--Tactile Representation Analysis}
\label{sec:supp_stage1}

We evaluate shared contact semantics, physical information retention, and missing-input robustness using the frozen joint wrist--tactile encoder employed by the task policies.

\subsection{Evaluation Samples and Semantic Readouts}

\paragraph{Held-out samples.}
We sample reference frames uniformly from held-out source episodes.
The paired semantic/probe subset includes only frames for which the required local-force, intensity, force/torque, contact, and strength targets pass their validity masks.
Metrics are pooled over eligible samples.
Training-fitted normalization and label calibration remain fixed throughout evaluation.

\paragraph{Common-head protocol.}
The tactile and wrist semantic projections are each passed to the same semantic head trained during joint representation learning.
A separate trained head reads the joint vector; it is also used when one modality is withheld.
All heads remain frozen during evaluation.
Contact accuracy thresholds the predicted probability at 0.5, and strength accuracy uses the most probable of the four classes.
Contact refers to tactile-pad contact with the grasped object; its relationship to strength labels is specified in Sec.~\ref{sec:supp_label_calibration}.

\paragraph{Probability quality.}
Table~\ref{tab:supp_semantic_probabilities} reports contact binary cross-entropy (BCE), contact Brier score (mean squared probability error), and strength cross-entropy (CE).
The contact-prior baseline predicts a constant probability estimated from training observations and obtains 81.37\% contact accuracy on the evaluation subset.
Its BCE and Brier score are 0.4830 and 0.1522.
The wrist readout improves on this contact prior, while retaining larger probability errors than the tactile readout.

\begin{table}[ht]
    \centering
    \caption{\textbf{Additional semantic readout metrics.} All entries use the same paired evaluation samples; lower is better.}
    \label{tab:supp_semantic_probabilities}
    \begin{tabular}{@{}lrrr@{}}
        \toprule
        Readout / available input & Contact BCE & Contact Brier & Strength CE \\
        \midrule
        Shared head / tactile projection & 0.0045 & 0.0008 & 0.2048 \\
        Shared head / wrist projection & 0.3436 & 0.0646 & 1.2108 \\
        Joint head / both modalities & 0.0484 & 0.0087 & 0.4841 \\
        Joint head / touch only & 0.0047 & 0.0008 & 0.2174 \\
        Joint head / wrist only & 0.4481 & 0.0672 & 1.4161 \\
        \bottomrule
    \end{tabular}
\end{table}

\subsection{Physical Information Retention}

\paragraph{Matched reconstruction probes.}
Figure~\ref{fig:stage1_threeway_main} compares force-intensity reconstructions from the frozen 64-dimensional tactile, wrist, and joint vectors.
The target contains two $7\times4$ maps: each cell averages native force-vector magnitudes before division by the fixed training RMS scale.
All three probes use exactly the same target and sample partition.

Paired samples are divided by source episode into disjoint probe-fitting and evaluation subsets using seed 42.
Both subsets use episodes held out from encoder training.
For each representation, features are standardized using only its probe-fitting subset, with each standard deviation floored at $10^{-5}$.
A ridge decoder predicts the flattened 56-cell target with squared error averaged over fitting samples and an L2 coefficient of 0.1.
The target mean from the fitting subset provides the intercept.
Evaluation MAE averages absolute error over frames, pads, and cells without smoothing or clipping predictions.

The tactile, wrist, and joint probe MAEs are 0.1557, 0.3744, and 0.1768, respectively.
The training-derived task-mean map has MAE 0.4850; shuffling joint vectors within task gives 0.5815.
The joint vector retains sample-specific physical information beyond task identity, while tactile-only encoding gives the lowest error on this tactile target.

\paragraph{Additional readouts.}
The native force-vector heads achieve normalized MAE of 0.0328 from tactile vectors and 0.0451 from joint vectors.
These heads predict signed three-component vectors and are scored using the corresponding valid-target masks.
The joint vector's visual-summary head has MAE 0.2574 against its frozen visual-feature target.

\paragraph{Reconstruction visualization.}
For insertion and wiping, the displayed example is the median-error contact frame under the joint probe within that task's probe-evaluation subset.
The same frame is shown for all three representations, using a shared color scale.
Linear predictions are displayed without clipping.

\subsection{Temporal Semantic Example}
\begin{figure}[ht]
    \centering
    \includegraphics[width=0.95\linewidth]{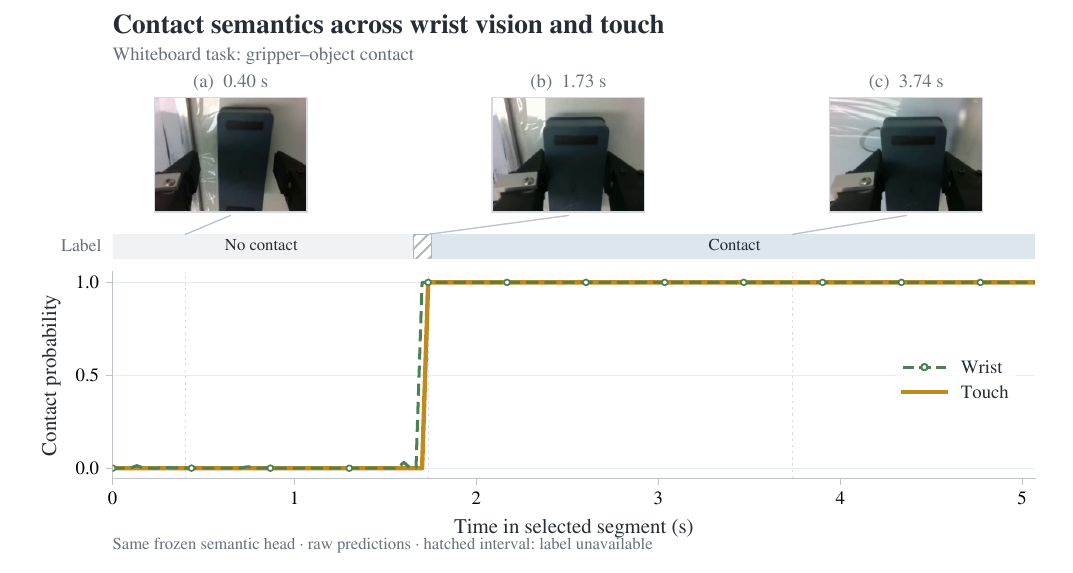}
    \caption{\textbf{Cross-modal contact semantics over time.} Wrist keyframes align with wrist and tactile contact probabilities. Contact denotes tactile-pad contact with the grasped object; hatched intervals have no valid contact label. Both curves use the same frozen semantic head and retain raw per-frame predictions.}
    \label{fig:contact_semantics_temporal}
\end{figure}

Figure~\ref{fig:contact_semantics_temporal} uses a held-out wiping sequence and the same trained semantic head on frozen wrist and tactile projections.
The displayed window covers the first valid contact transition in the sequence.
Wrist frames and raw contact probabilities are aligned by their recorded timestamps.
Unavailable contact labels remain explicitly marked.

\subsection{Single-Pad and Missing-Modality Diagnostics}

We evaluate seven local-input patterns on the same paired samples using the frozen joint semantic head.
Inputs are withheld before encoding; target labels and validity masks remain fixed.

\begin{table}[ht]
    \centering
    \caption{\textbf{Offline availability evaluation.} Joint-head contact and strength accuracies (\%) on matched samples.}
    \label{tab:supp_pad_audit}
    \begin{tabular}{@{}lrr@{}}
        \toprule
        Available local inputs & Contact & Strength \\
        \midrule
        Both pads and wrist & 99.06 & 91.37 \\
        Both pads only & 99.91 & 93.25 \\
        Wrist only & 92.22 & 78.97 \\
        Right pad and wrist (left absent) & 92.91 & 83.93 \\
        Left pad and wrist (right absent) & 94.53 & 84.44 \\
        Left pad only & 95.56 & 84.02 \\
        Right pad only & 87.69 & 85.04 \\
        \bottomrule
    \end{tabular}
\end{table}

Contact semantics remain readable with one pad absent, with performance varying by pad and input combination.
When all local inputs are absent, the encoder returns an invalid placeholder, which is excluded from semantic scoring.

\section{Autonomous Recovery}
\label{sec:supp_recovery}

\subsection{Failure Confirmation}

A transient force peak or motion pause alone does not establish failure.
For example, both successful seating and obstruction against the socket panel can produce substantial resistance during insertion.
Failure confirmation therefore requires the preceding force and motion history to determine whether the interaction is making effective progress.

We deploy a lightweight temporal detector on the robot-side controller, using bilateral tactile force/torque and joint/gripper motion history.
A single-layer, unidirectional GRU with 128 hidden units and a lightweight classification head estimates a failure score from current and past observations.
When this score remains above the task-specific calibrated threshold for a sustained interval under valid, fresh observations, the detector confirms failure and initiates correction.
Temporal confirmation suppresses triggers caused by brief force fluctuations or pauses during normal manipulation.

The detector is trained independently of the action policy using task-specific contact and motion event annotations and binary cross-entropy.
Uncertain intervals are excluded from supervision, and training, calibration, and evaluation are separated by episode.

\subsection{Correction and Resuming Execution}

Once failure is confirmed, the system records an interaction snapshot and supplies it, together with an adjustment instruction, as an additional condition for the Action expert.
During correction, the failure snapshot remains fixed while the current wrist--tactile representation continues to update.
The policy combines this failure context with the evolving interaction state to generate withdrawal, repositioning, or contact-restoration actions learned from corrective demonstrations.

Correction ends when the detector confirms recovery completion.
For insertion, a separate completion head shares the GRU state and uses changes relative to the failure snapshot to identify withdrawal and reduced obstruction load while the plug remains held.
This head is trained with binary cross-entropy while the failure branch remains fixed.
Completion evidence must persist and remain valid at an action-chunk boundary after at least 32 corrective action steps have executed.
For wiping, the detector uses contact and motion evidence to confirm that effective eraser--surface contact has been restored during wiping motion.

After recovery confirmation, the adjustment instruction and additional failure condition are disabled, and normal task execution resumes.
A subsequent failure may trigger another correction within the trial budget.

\subsection{Validation}

\paragraph{Offline event agreement.}
For insertion, causal replay against rule-derived event labels gives 95.0\% failure recall.
Recovery-completion agreement is 94.7\% among attempts with an eligible completion boundary.
The combined failure-detection and recovery-completion rate is 86.3\% over all failed attempts.

\paragraph{Real-robot task completion.}
Under the retry protocol in Sec.~\ref{sec:supp_protocol}, autonomous retries raise insertion success from 50\% to 80\% and wiping success from 80\% to 100\%.
Each condition uses 20 complete-task trials; final success includes both initial-attempt successes and successes after correction.

\end{document}